\documentclass[]{interact}

\usepackage{epstopdf}
\usepackage{subfigure}

\usepackage{natbib}
\bibpunct[, ]{(}{)}{;}{a}{}{,}

\theoremstyle{plain}

\theoremstyle{definition}

\theoremstyle{remark}

\usepackage{epsfig,comment,cite,latexsym,color,balance,hyperref}

\newcommand{\beqn}{\begin{eqnarray}}
\newcommand{\eeqn}{\end{eqnarray}}
\newcommand{\beq}{\begin{eqnarray*}}
\newcommand{\eeq}{\end{eqnarray*}}
\newcommand{\e}{\mbox{{\boldmath${e}$\unboldmath}}}  

\newcommand{\C}{\mbox{{\boldmath${C}$\unboldmath}}}  

\title{A version of the Cloude/van Zyl decomposition for full/quad pol synthetic aperture radar data}
\author{
\name{Allan Aasbjerg Nielsen\textsuperscript{a}\thanks{CONTACT Allan A.\ Nielsen. Email: alan@dtu.dk. ORCID 0000-0002-4837-9449.}, Paul Jacques Connetable\textsuperscript{b}, Henning Skriver\textsuperscript{b} and\\Knut Conradsen\textsuperscript{a}}
\affil{\textsuperscript{a}DTU Compute -- Applied Mathematics and Computer Science, Building 324\\\textsuperscript{b}DTU Space -- National Space Institute, Building 348\\Technical University of Denmark, DK-2800 Lyngby, Denmark}
}

\begin{document}

\maketitle


\thispagestyle{empty}

\urlstyle{same}

\begin{abstract}
Based on the Cloude/van Zyl decomposition of reflection symmetric polarimetric SAR data
we present a version for full/quad polarisation data.
We demonstrate both versions on X-band F-SAR data over Vejers, Denmark.
Differences between the original and our version of the decomposition occur especially for very bright man-made targets.
These differences come with lower entropy for our version giving an improved focus on a more pronounced dominant scattering mechanism for these objects.
Interpretation of the scattering mechanism related to the
full/quad pol version of the 
decomposition is facilitated by means of the
Cloude/Pottier 
$\bar{\alpha}$
angle.
\end{abstract}

\begin{keywords}
Complex covariance matrices,
Hermitian matrices,
polarimetric SAR.
\end{keywords}

\section{Introduction}
\label{sec:intro}

In
\citep{vanzyl1992}
van Zyl describes what he calls the Cloude decomposition
\citep{cloude1988}
for polarimetric synthetic aperture radar (SAR) data.
The Cloude decomposition is the general spectral decomposition
consisting of sums of products of eigenvalues (sorted non-increasingly) and outer products of the corresponding eigenvectors of the 
Hermitian 3$\times$3 covariance matrix of multilook SAR data.
In
\citep{vanzyletal2011}
which deals primarily with another subject, namely an improvement of the Freeman/Durden decomposition
\citep{freemandurden1998},
the decomposition in
\citep{vanzyl1992}
for reflection symmetric data is described further with an explicit distinction between single and double bounce
(dihedral)
targets.
Inspired by this work we here give a version of the decomposition for full/quad pol SAR data.

For both the original, reflection symmetry and the full/quad pol versions of the decomposition,
the eigenvalues needed are calculated by means of fast methods described in
\citep{uspensky1948,fastLoewner2019}.

Shortcomings in the van Zyl decomposition relating to asymmetric scattering are dealt with in
\citep{BhattaTouzi2011},
and the single/double bounce dichotomy for full pol covariance matrices in
\citep{Cui_et_al2014}.
In this paper these issues are not addressed further.

The data used here to illustrate the decompositions are X-band data from the
F-SAR system
\citep{FSAR2013}
from the German Aerospace Center (DLR)
covering a small area in western Denmark.

\section{Multilook polarimetric SAR data}
\label{sec:polsar}

In the covariance matrix formulation of multilook  polarimetric SAR (polSAR) image data each pixel is described by a complex 3$\times$3 matrix
\beq
\label{eq:covmatgreek}
\C & = &
\left[ \begin{array}{rrr}
\langle S_{hh}S_{hh}^* \rangle & \sqrt{2}\langle S_{hh}S_{hv}^* \rangle & \langle S_{hh}S_{vv}^* \rangle \\
\sqrt{2}\langle S_{hv}S_{hh}^* \rangle & 2\langle S_{hv}S_{hv}^* \rangle & \sqrt{2}\langle S_{hv}S_{vv}^* \rangle \\
\langle S_{vv}S_{hh}^* \rangle & \sqrt{2}\langle S_{vv}S_{hv}^* \rangle & \langle S_{vv}S_{vv}^* \rangle \\
\end{array} \right]
.
\eeq
This matrix is Hermitian also known as self-adjoint, i.e., the matrix is equal to its own conjugate transpose,
$\C = \C ^H$,
the superscript $^H$ denotes transpose and complex conjugate (which is denoted by $^*$),
$\langle \rangle$ denotes ensemble averaging.
The trace, the determinant and all eigenvalues are real and nonnegative.

The general spectral decomposition of this matrix is
\beq
\C &=& \lambda_1 \e_1\e_1^H + \lambda_2 \e_2\e_2^H + \lambda_3 \e_3\e_3^H
\label{eq:C}
\eeq
where $\lambda_i$ are the eigenvalues of \C\ and $\e_i$ the corresponding mutually orthogonal eigenvectors.

\section{Eigenvalues and -vectors}
\label{sec:eig}

In this section we solve the eigenvalue problem for the reflection symmetry case.
For (fast) calculation of eigenvalues in the full/quad pol case we refer to
\citep{uspensky1948,fastLoewner2019,cloude1997,jsleeetal1999}.

\subsection{Reflection Symmetry}
\label{sec:refl}

With the notation from
\citep{vanzyl1992,vanzyletal2011}
the reflection symmetry version of the covariance matrix is
\beq
\C_r &=& 
\left[ \begin{array}{ccc}
\xi      & 0      & \rho  \\
0     & \eta  & 0     \\
\rho^*  & 0  & \zeta \\
\end{array} \right]
.
\label{eq:Caz}
\eeq
The eigenvalues $\lambda_i, i=1,2,3$ are found by solving
\beq
(\xi-\lambda_i)(\eta-\lambda_i)(\zeta-\lambda_i)-(\eta-\lambda_i)|\rho|^2 &=& 0
\eeq
leading to
\beq
\lambda_{1,2} &=&
\frac{\zeta+\xi \pm \sqrt{\Delta}}{2}\\
\lambda_3 &=& \eta
,
\eeq
where
$\Delta = (\zeta-\xi)^2+4|\rho|^2$ ($\Delta \geq 0$) and
$\lambda_1 \geq \lambda_2$.

The first and second eigenvalues are both
associated with either single bounce scattering (odd number of reflections) or double bounce scattering (even number of reflections),
see~(\ref{eq:Cazdecomp}).
The third eigenvalue is the covariance matrix element
$\eta = 2\langle S_{hv}S_{hv}^* \rangle$
associated with the remainder, often volume scattering in vegetation
or dihedrals rotated with respect to the flight line.
Circular polarization to avoid the rotation dependency and determination of the rotation angle are described in
\citep{jslee_etal2002,yamaguchi_etal2011}.
In
\citep{yamaguchi_etal2011}
the angle is determined by cross-polarization minimization.
%
Neither issue is pursued further here.

With
\beq
\Lambda_1 &=& \lambda_1 \, \frac{[\zeta-\xi+\sqrt{\Delta}]^2}{[\zeta-\xi+\sqrt{\Delta}]^2+4|\rho|^2} \\
\Lambda_2 &=& \lambda_2 \, \frac{[\zeta-\xi-\sqrt{\Delta}]^2}{[\zeta-\xi-\sqrt{\Delta}]^2+4|\rho|^2}\\
\Lambda_3 &=& \lambda_3\\
\alpha_{v} &=& \frac{2\rho}{\zeta-\xi+\sqrt{\Delta}}\\
\beta_{v} &=& \frac{2\rho}{\zeta-\xi-\sqrt{\Delta}}
,
\eeq
we have the corresponding eigenvectors, see
\citep{vanzyl1992,vanzyletal2011}
\beq
\e_1 &=& \sqrt{\frac{[\zeta-\xi+\sqrt{\Delta}]^2}{[\zeta-\xi+\sqrt{\Delta}]^2+4|\rho|^2}}
\left[ \begin{array}{c}
\alpha_{v}\\
0\\
1\\
\end{array} \right]\\
\e_2 &=& \sqrt{\frac{[\zeta-\xi-\sqrt{\Delta}]^2}{[\zeta-\xi-\sqrt{\Delta}]^2+4|\rho|^2}}
\left[ \begin{array}{c}
\beta_{v}\\
0\\
1\\
\end{array} \right]\\
\e_3 &=& 
\left[ \begin{array}{c}
0\\
1\\
0\\
\end{array} \right]
.
\eeq
For $\e_3$ we could choose any complex vector with first and third components equal to zero and second component on the unit circle in the complex plane.
As in
\citep{vanzyl1992,vanzyletal2011}
we choose $1$.
Incidentally, each $\e_i$ has arbitrary complex phase $\exp(j\phi_i)$.

For $|\rho| > 0$ the denominator of $\alpha_{v}$ is positive and
the denominator of $\beta_{v}$ is negative.
Also,
$\alpha_{v} \beta_{v}^* = \alpha_{v}^* \beta_{v} = -1$
(leading to $\e_1^H \e_2 = 0$; trivially $\e_1^H \e_3 = \e_2^H \e_3 = 0$)
and since for any complex number $z$, $1/z = z^*/|z|^2$ we have the relations
\beq
\beta_{v} = -\frac{1}{\alpha_{v}^*} = -\frac{\alpha_{v}}{|\alpha_{v}|^2} \mbox{ \ \ \ and \ \ \ } |\beta_{v}|^2 = \frac{1}{|\alpha_{v}|^2}
.
\eeq

This leads to the decomposition
\beqn
\label{eq:Cazdecomp}
\nonumber
\C_r
&=& \lambda_1 \e_1\e_1^H + \lambda_2 \e_2\e_2^H + \lambda_3 \e_3\e_3^H\\ \nonumber
%
%
&=&
\Lambda_1
\left[ \begin{array}{ccc}
|\alpha_{v}|^2  & 0 & \alpha_{v}\\
0 & 0  & 0 \\
\alpha_{v}^*  & 0  & 1\\
\end{array} \right]
+
\Lambda_2
\left[ \begin{array}{ccc}
|\beta_{v}|^2 & 0 & \beta_{v}\\
0 & 0 & 0\\
\beta_{v}^* & 0 & 1\\
\end{array} \right]
+\\
&&
\Lambda_3
\left[ \begin{array}{ccc}
0 & 0 & 0\\
0 & 1 & 0\\
0 & 0 & 0\\
\end{array} \right]
.
\eeqn

To explicitly distinguish between single and double bounce scattering,
we sort the above parameters such that
\begin{itemize}
\item
$\lambda_1$, $\e_1$, $\Lambda_1$ and $\alpha_{v}$ more closely represent single bounce scattering
(the argument of
$\rho = \langle S_{hh}S_{vv}^* \rangle$
is closer to $0$ than to $\pm \pi$
 which means that
 the real part of
$\rho$
is nonnegative),
and
\item
$\lambda_2$, $\e_2$, $\Lambda_2$, and $\beta_{v}$ more closely represent double bounce scattering
(the argument of
$\rho = \langle S_{hh}S_{vv}^* \rangle$
is closer to $\pm \pi$ than to $0$
 which means that
 the real part of
$\rho$
is negative).
\end{itemize}

As a measure of target disorder, entropy $H$ is defined as
\beq
H = -\sum_{i=1}^3 p_i \log_3 p_i
\mbox{ \ \ \ with \ \ \ }
p_i = \frac{\lambda_i}{\sum_{i=1}^3 \lambda_i},
\eeq
$0 \leq H \leq 1$. 
Low entropy is associated with a single dominant scattering mechanism, high entropy with random scattering.

A good RGB visualization of the decomposition is
\begin{itemize}
\item red: double bounce scattering given by\\$\lambda_2 = \Lambda_2(|\beta_{v}|^2+1)$; $|\beta_{v}|^2+1$ is the 
trace of the second matrix in the decomposition of  $\C_r$ in~(\ref{eq:Cazdecomp}),
\item green: volume scattering given by $\lambda_3 = \Lambda_3$,
\item blue: single bounce scattering given by\\$\lambda_1 = \Lambda_1(|\alpha_{v}|^2+1)$; $|\alpha_{v}|^2+1$ is the 
trace of the first matrix in the decomposition of  $\C_r$ in~(\ref{eq:Cazdecomp}).
\end{itemize}


If one is less interested in the strongly reflecting man-made objects in the scene but more interested in overall structure, the RGB visualization can be made by means of square roots or logarithms of the eigenvalues.
This will enhance the visual impression of the weaker reflectors.

\subsection{Full/Quad Pol}
\label{sec:fullpol}

For full/quad pol we have the full covariance matrix
\beq
\C &=& 
\left[ \begin{array}{ccc}
\xi      & a      & \rho  \\
a^*     & \eta  & b     \\
\rho^*  & b^*  & \zeta \\
\end{array} \right]
\eeq
and the eigenvalues are found as described in
\citep{uspensky1948,fastLoewner2019}.
The solutions given in
\citep{fastLoewner2019}
are well suited for fast array based computer implementation.

In this full/quad pol case the third eigenvalue, $\lambda_3$, is the one closest to 
$\eta = 2\langle S_{hv}S_{hv}^* \rangle$.
Again,
we sort the remaining two eigenvalues $\lambda_1$ and $\lambda_2$ according to the sign of the real part of
$\rho = \langle S_{hh}S_{vv}^* \rangle$,
i.e., according to whether the argument of
$\rho$
is closer to $0$ or to $\pm \pi$.

One could argue that in the full/quad pol case unlike in the reflection symmetry case, the eigenvectors do not have direct interpretations as being associated with specific scattering mechanisms (single bounce, double bounce, volume scattering).
On the other hand
the Cloude/van Zyl decomposition for reflection symmetric data places restrictions on the data \mbox{\em before} the decomposition---especially, man-made objects  certainly do not exhibit reflection symmetry---whereas what we do here challenges the interpretation \mbox{\em after} the decomposition.

To facilitate the interpretation of the scattering mechanism in this full matrix case we use the
$\bar{\alpha}$ angle suggested by Cloude and Pottier
\citep{cloude1997,jsleeetal1999}.
$\bar{\alpha}$ is calculated from the eigenvectors of the coherency matrix
%
(we reuse the symbol $\e_i, i=1,2,3$ for eigenvectors)
\beq
\e_i &=&
\exp(j\phi_i)
\left[ \begin{array}{c}
\cos \alpha_i\\
\sin \alpha_i \cos \beta_i \exp(j\delta_i)\\
\sin \alpha_i \sin \beta_i \exp(j\gamma_i)\\
\end{array} \right]
\eeq
where $\alpha_i$, $\beta_i$,
$\delta_i$ and $\gamma_i$ are angles which characterize the scattering.
Here we are interested in the $\alpha_i$ only.
For the first element $e_{i1}$ of $\e_i$ we have
\beq
|e_{i1}|^2
&=& e_{i1} e_{i1}^*\\
&=& \exp(j\phi_i) \cos \alpha_i  \exp(-j\phi_i) \cos \alpha_i\\
&=& \cos^2 \alpha_i
\eeq
and therefore
\beq
\alpha_i &=& \arccos |e_{i1}|
\eeq
($\arccos$ is sometimes written as $\cos^{-1}$).
%
For $\bar{\alpha}$ we have
\beq
\bar{\alpha} &=& \sum_{i=1}^3 p_i \alpha_i
\eeq
with the $p_i$ calculated as in the above subsection
\ref{sec:refl}
but now based on the eigenvalues of the full matrix
(the full covariance matrix and the coherency matrix have the same eigenvalues).
%
%
Alternatively,
the $\alpha_i$
may be calculated by means of the
eigenvector-eigenvalue identity
where we obtain the $|e_{i1}|^2$ without calculating
$\e_i$
\citep{hensley2021,denton_etal2022,EEI2022}.
This speeds up calculations immensely compared to calls to built-in eigensolvers in for-loops over rows an columns, see
\citep{EEI2022}.
For $\bar{\alpha}$ we have
\begin{itemize}
\item
$\bar{\alpha}=0^{\circ}$ is associated with single bounce (surface) scattering, 
\item
$\bar{\alpha}=45^{\circ}$ with scattering from differently oriented dipoles
(typically vegetation
or dihedral targets rotated with respect to the flight line),
and
\item
$\bar{\alpha}=90^{\circ}$ with (conductive) double bounce scattering.
\end{itemize}


\section{Data example}

In a project headed by the
Danish Ministry of Defence Acquisition and Logistics Organization
(DALO)  the F-SAR
system of the German Aerospace Center (DLR) has been used to acquire data at several locations in Denmark and Greenland.
The F-SAR offers
the possibility of exploring the performance of a high resolution,
fully polarimetric SAR system with five frequency
bands in the range $\sim$300 MHz to $\sim$10 GHz
\citep{FSAR2013}.

\begin{figure} [t!]
\centering
\includegraphics[trim={0cm 0cm 0cm 0cm},clip,width=\linewidth]{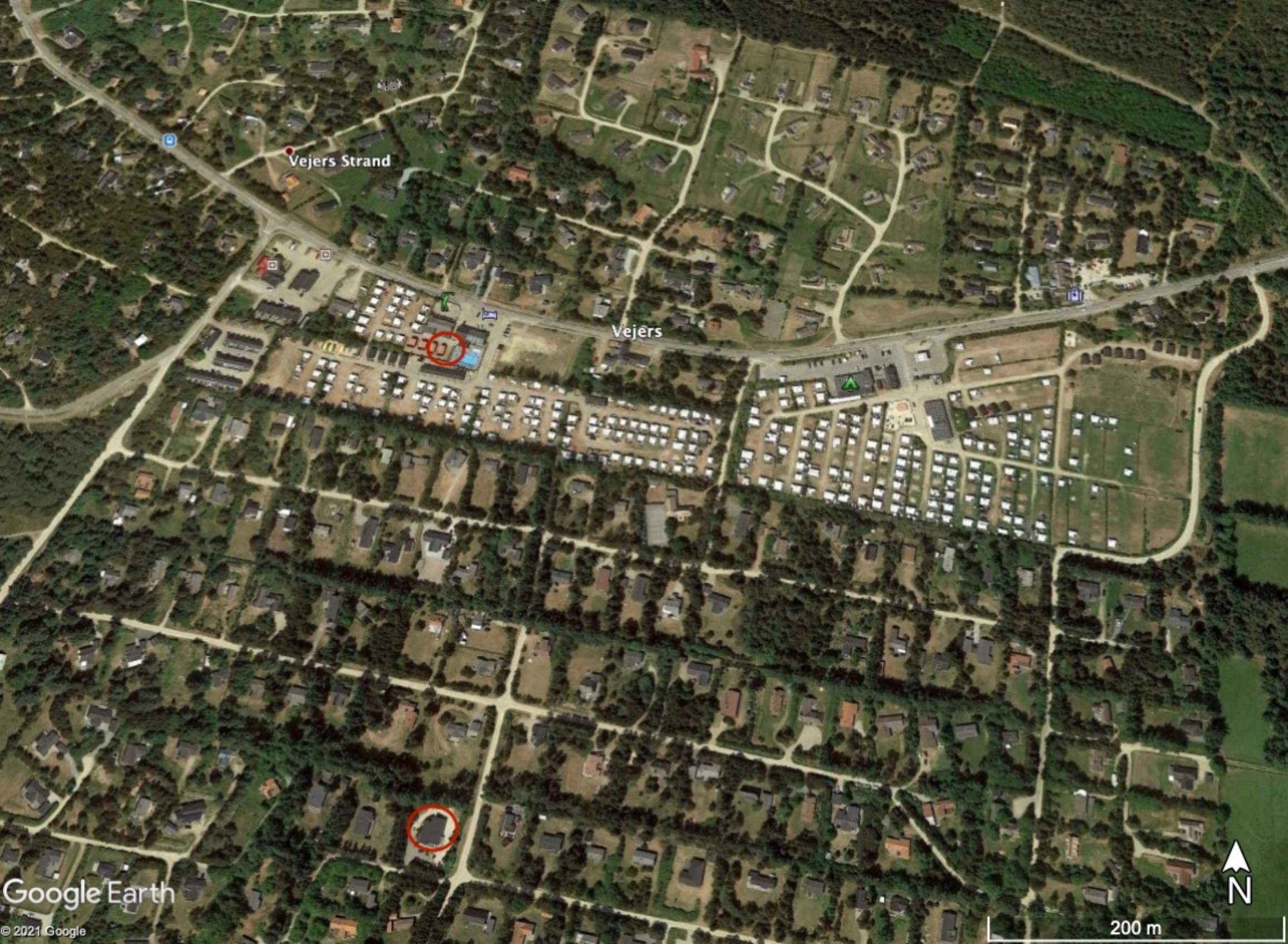}
\caption{\label{fig:vejers}
Part of Vejers, Denmark, approximately the area covered by the F-SAR data, from Google Earth.
We see summer houses, caravan sites with a swimming pool and service buildings (a grocery store, a petrol station etc.), the main road called Vejers Havvej (where the text ``Vejers'' appears) going mostly east-west, dirt roads, vegetation etc.
Note the two red circles in the left part of the image, top-center (near a swimming pool) and bottom (a big house).}
\end{figure}

\begin{figure} [t!]
\centering
\includegraphics[width=\linewidth]{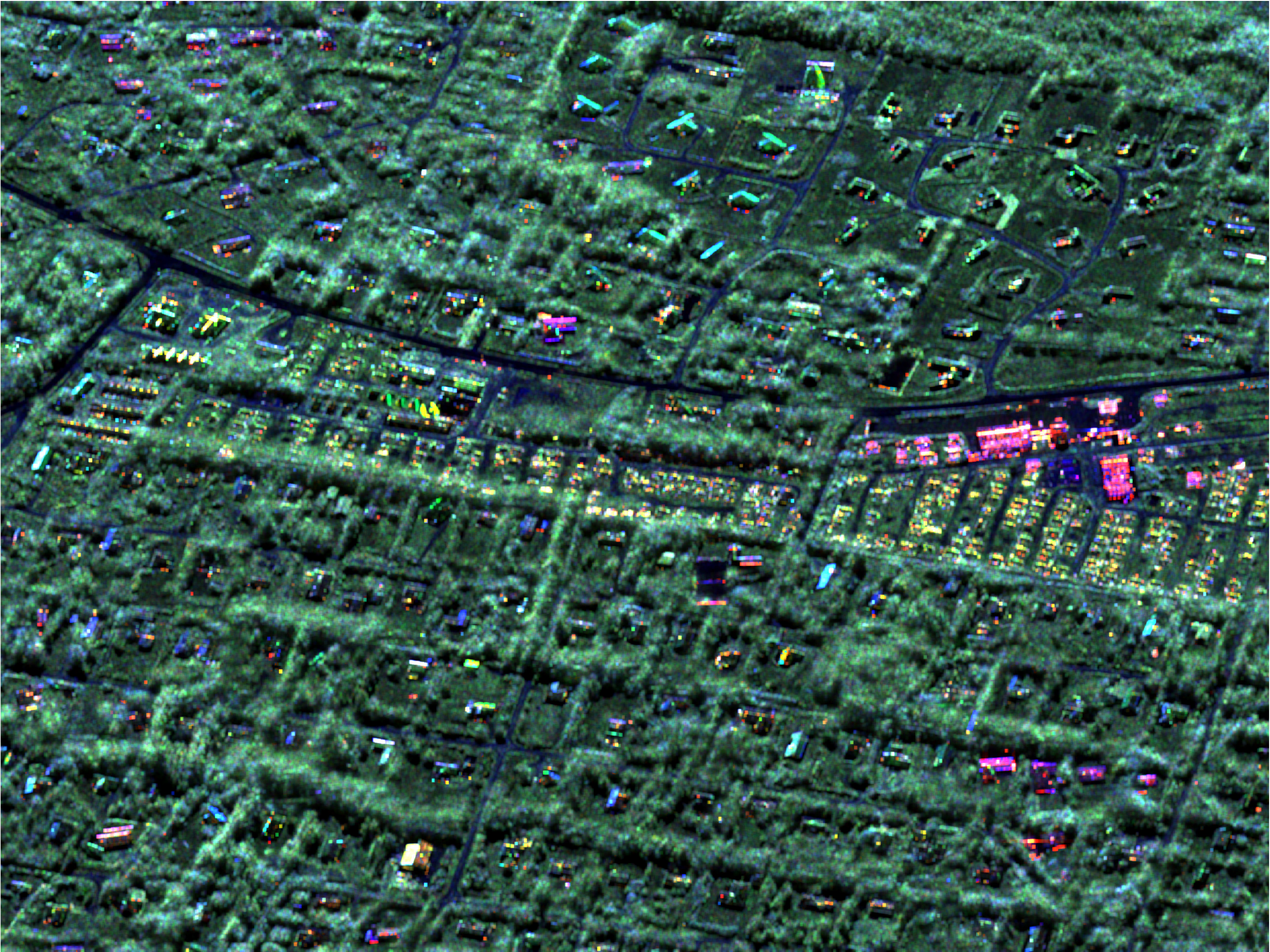}
\caption{\label{fig:vzazim}
Original reflection symmetry based Cloude/van Zyl decomposition.
Red is $\sqrt{\lambda_2}$ (double bounce scattering or even number of reflections), green is $\sqrt{\lambda_3}$ (volume scattering), blue is $\sqrt{\lambda_1}$ (single bounce scattering or odd number of reflections).}
\end{figure}

\begin{figure} [t!]
\centering
\includegraphics[width=\linewidth]{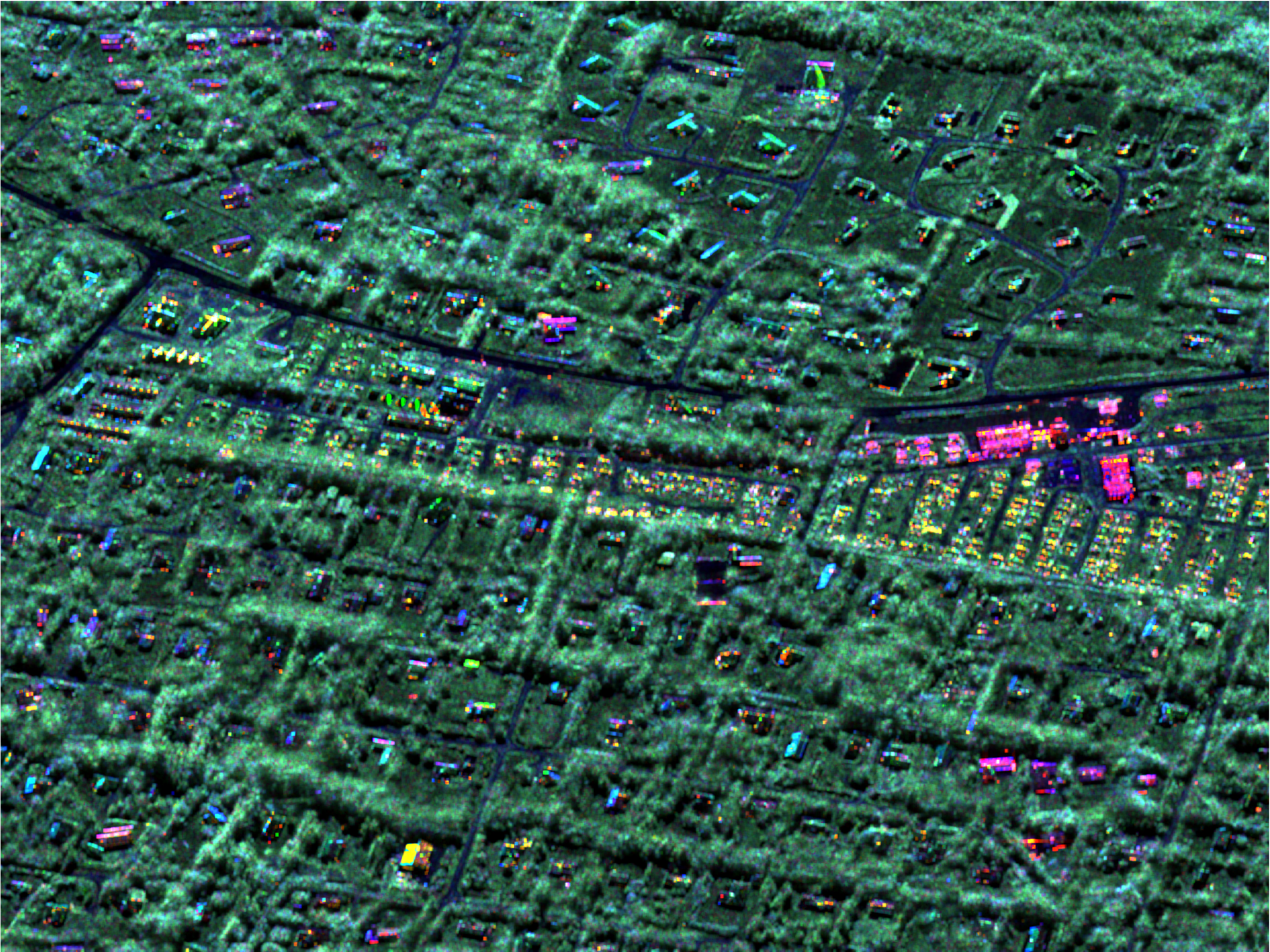}
\caption{\label{fig:vzfull}
Full/quad pol version of Cloude/van Zyl decomposition.
Red is $\sqrt{\lambda_2}$, green is $\sqrt{\lambda_3}$, blue is $\sqrt{\lambda_1}$.
A few differences from Figure~\ref{fig:vzazim} can be seen for some strong scatterers, see text and Figure~\ref{fig:diffentropy}.}
\end{figure}

The decompositions are illustrated 
with airborne F-SAR X-band data (frequency 9.60 GHz corresponding to a wavelength of $\sim$3 cm).
The data were acquired on 20 June 2017
covering a central part of the village Vejers on the Danish west coast, Figure~\ref{fig:vejers}.
The scene covers the east-west main road (called Vejers Havvej where the text ``Vejers'' appears in Figure~\ref{fig:vejers} and depicted in black in Figures~\ref{fig:vzazim} and \ref{fig:vzfull}) going to and from the sea, camp sites with caravans to the south of the road, a swimming pool, parking lots and several service buildings (grocery store, petrol station etc.)\ as well as many small and big houses north and south of the road.
Note the two red circles.

\begin{figure} 
\centering
\includegraphics[trim={0cm 0cm 0cm 0cm},clip,width=\linewidth]{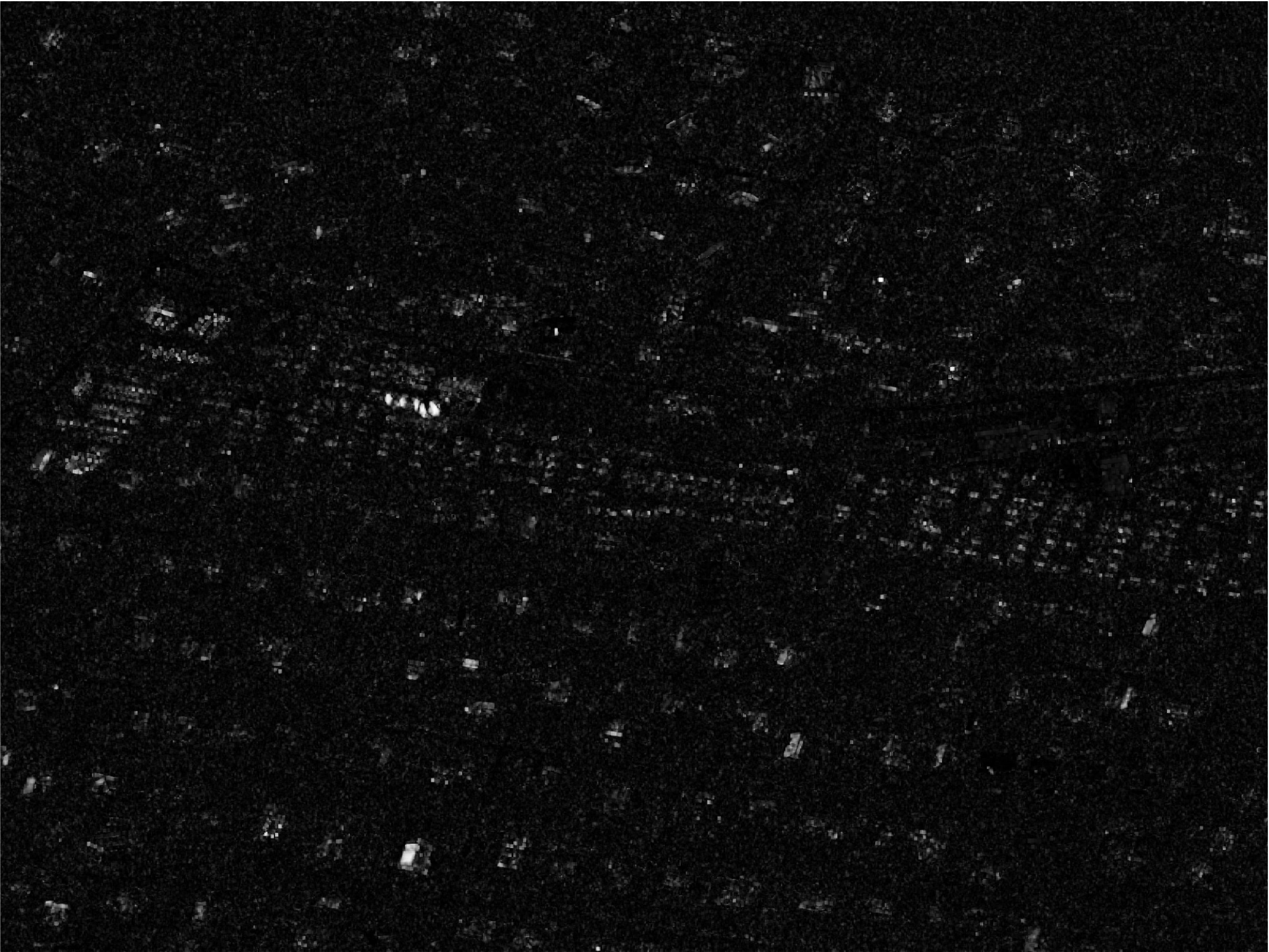}
\caption{\label{fig:diffentropy}
Entropy (in the interval $[0,1]$) for the original reflection symmetry decomposition minus entropy ($[0,1]$) for the full/quad pol version.
Differences are consistently nonnegative, i.e, we always have lower entropy for the full/quad pol version.
Large differences (up to $0.62$) are concentrated on bright man-made objects (houses, caravans etc.).
Hence the suggested version gives less target scattering mechanism disorder for these objects.}
\end{figure}

In Figures~\ref{fig:vzazim} and \ref{fig:vzfull}
in which square roots of the eigenvalues are shown as RGB,
red objects
are mainly buildings which are characterized by double bounce reflection.
Green objects in the two images are mainly
characterized by volume scattering typically vegetation
or dihedral targets which are rotated with respect to the flight line.
Blue areas in the two images are objects which are characterized by single bounce reflection.
Magenta is a mix of red and blue and therefore magenta objects represent a combination of double and single bounce reflection.
Yellow is a mix of red and green and therefore yellow objects represent a combination of double  bounce reflection and volume scattering.
Cyan is a mix of green and blue and therefore cyan objects represent a combination of volume 
scattering and single bounce reflection.
White and gray objects represent
a combination of
all three types of scattering.

In a few places differences between the two decompositions can be seen (two examples are marked by red circles in Figure~\ref{fig:vejers}).
For example a large whitish trapezoid-shaped roof of the circled building to the bottom-left in Figure~\ref{fig:vzazim}
appears more yellow in Figure~\ref{fig:vzfull}.
This means that our version of the decomposition suggests two scattering mechanisms as opposed to three for the original one.
Also, two yellow building parts (also circled) just south of Vejers Havvej in the middle-left part of Figure~\ref{fig:vzazim}
immediately left of the swimming pool seen in Figure~\ref{fig:vejers}
are red in Figure~\ref{fig:vzfull}.
This means that our version of the decomposition suggests one scattering mechanism as opposed to two for the original one.
Similar considerations are valid also for other targets not mentioned here.

Figure~\ref{fig:diffentropy}
shows the difference between entropies
as obtained by the original and the full matrix version of the decomposition.
Differences are consistently nonnegative
(the minimum value is 0.00, the maximum value is 0.62)
and large differences are on bright man-made objects (houses, caravans etc.).
This means that the suggested full matrix
version gives less
target scattering mechanism disorder, i.e., better scattering description for these objects.

\begin{figure} 
\centering
\includegraphics[trim={0cm 0cm 0cm 0cm},clip,width=0.8\columnwidth]{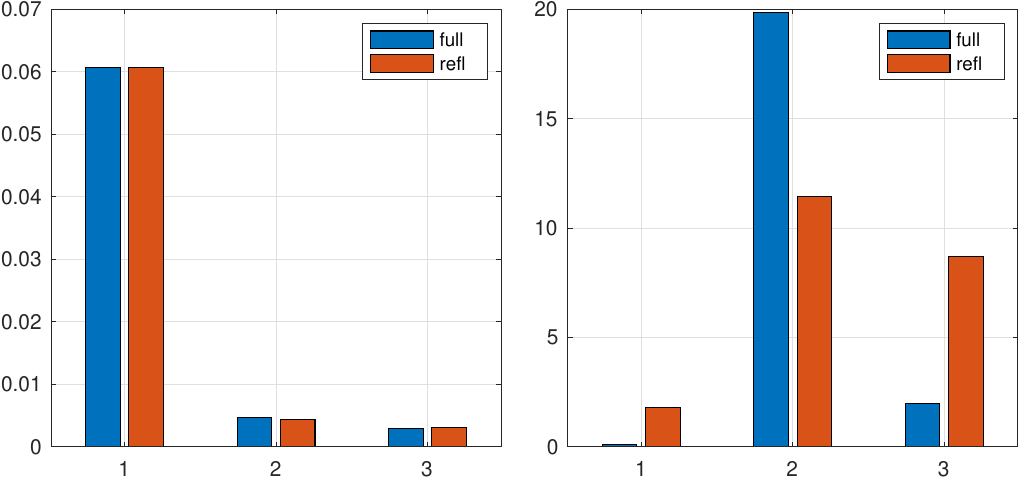}
\caption{\label{fig:vzhisto}
Eigenvalues for the Cloude/van Zyl decomposition, original version for reflection symmetry in red, our version for the full covariance matrix in blue.
For the original version ``1'' represents single bounce, ``2'' double bounce and ``3'' volume scattering.
The left figure shows eigenvalues for a (very low backscatter) road pixel where the two sets of eigenvalues are very similar.
Contrary to this the right figure shows eigenvalues for a (high backscatter) pixel on the roof of the circled house where the two sets are very different.
Note the very different scales on the two y-axes.}
\end{figure}

Figure~\ref{fig:vzhisto}
shows
eigenvalues for the two versions of the van Zyl decomposition
for two different situations: a road pixel (left) and a pixel on the roof of the circled house (right),
original version for reflection symmetry in red, our version for the full covariance matrix in blue.
For the road the eigenvalues are very similar.
For the house, unlike the case for the original decomposition, we get one dominant scattering mechanism and hence less scattering mechanism disorder for our version.

Table~\ref{tab:alpha}
shows
$\bar{\alpha}$
for a few selected pixels in Figure~\ref{fig:vzfull} including the two circled targets in Figure~\ref{fig:vejers}.
Correspondence with the expected values for different scattering mechanisms mentioned in Section~\ref{sec:fullpol} is satisfactory.
Specifically, we see that the circled big house for our version exhibits the expected double bounce scattering.

\begin{table}
\centering
\caption{\label{tab:alpha} $\bar{\alpha}$ for a few selected pixels in Figure~\ref{fig:vzfull}.}
\vspace{0.5\baselineskip}
\begin{tabular}{|l|c|} \hline\hline
& $\bar{\alpha}$\\ \hline
Big house (circled) & $68^{\circ}$\\ 
West of swimming pool  (circled) & $62^{\circ}$\\  
Wood & $30^{\circ}$\\ 
Road & $11^{\circ}$\\ \hline\hline
\end{tabular}
\end{table}


\section{Conclusions}

We propose
a version of the Cloude/van Zyl decomposition for full/quad pol synthetic aperture radar data.
The Cloude/van Zyl decomposition for reflection symmetric data
offers direct interpretation of the eigenvectors as being associated with single bounce, double bounce or volume scattering
but places restrictions on the data before the decomposition, especially, man-made objects  certainly do not exhibit reflection symmetry.
Contrary to this,
since the eigenvectors in the full/quad pol case do not have direct interpretations as being associated with the same specific scattering mechanisms,
what we do here challenges the interpretation after the decomposition.
To facilitate interpretation of the scattering mechanism in this case we use the $\bar{\alpha}$ angle.

The
explicit distinction between single and double bounce
targets in the original reflection symmetry Cloude/van Zyl decomposition
obtained by sorting the first two eigenvalues
according to whether the phase of $\langle S_{hh}S_{vv}^* \rangle$ is closer to $0$ or to $\pm \pi$,
is retained in this full/quad pol version.

In the case given,
differences between our version and the original decomposition based
%
%
on reflection symmetry occur especially for bright, man-made objects.
With our version these objects have consistently lower entropy.
Hence the suggested version gives less target scattering mechanism disorder for these objects.
%
%
This is depicted also in the visualizations shown, e.g., white objects for reflection symmetry to yellow objects for full/quad pol, or yellow to red objects.
%

Matlab code 
for calculating eigenvalues and entropy
for the two decompositions, and their visualization
is
available
on the
first
author's homepage https://people.compute.dtu.dk/alan/ under ``Publications''.
%





\bibliographystyle{chicago}
\bibliography{../wishart}

\begin{thebibliography}{}

\bibitem[\protect\citeauthoryear{Bhattacharaya and Touzi}{Bhattacharaya and
  Touzi}{2011}]{BhattaTouzi2011}
Bhattacharaya, A. and R.~Touzi (2011).
\newblock Polarimetric {SAR} urban classification using the {Touzi} target
  scattering decomposition.
\newblock {\em Canadian Journal of Remote Sensing\/}~{\em 37\/}(4), 323--332.
\newblock https://doi.org/10.5589/m11-042.

\bibitem[\protect\citeauthoryear{Cloude}{Cloude}{1992}]{cloude1988}
Cloude, S.~R. (1992).
\newblock Uniqueness of target decomposition theorems in radar polarimetry.
\newblock In {\em Direct and Inverse Methods in Radar Polarimetry, Part I, NATO
  Advanced Science Institutes Series}, pp.\  267--296.
\newblock (Event Bad Windsheim, Germany, 18-24 September 1988).

\bibitem[\protect\citeauthoryear{Cloude and Pottier}{Cloude and
  Pottier}{1997}]{cloude1997}
Cloude, S.~R. and E.~Pottier (1997, Jan.).
\newblock An entropy based classification scheme for land applications of
  polarimetric {SAR}.
\newblock {\em IEEE Transactions on Geoscience and Remote Sensing\/}~{\em
  35\/}(1), 68--78.
\newblock https://doi.org/10.1109/36.551935.

\bibitem[\protect\citeauthoryear{Cui, Yamaguchi, Yang, Kobayashi, S.-E, and
  Singh}{Cui et~al.}{2014}]{Cui_et_al2014}
Cui, Y., Y.~Yamaguchi, J.~Yang, H.~Kobayashi, P.~S.-E, and G.~Singh (2014,
  Apr.).
\newblock On complete model-based decomposition of polarimetric {SAR} coherency
  matrix data.
\newblock {\em IEEE Transactions on Geoscience and Remote Sensing\/}~{\em
  52\/}(4), 1991--2001.
\newblock https://doi.org/10.1109/TGRS.2013.2257603.

\bibitem[\protect\citeauthoryear{Denton, Parke, Tao, and Zhang}{Denton
  et~al.}{2022}]{denton_etal2022}
Denton, P.~B., S.~J. Parke, T.~Tao, and X.~Zhang (2022, Jan.).
\newblock Eigenvectors from eigenvalues: A survey of a basic identity in linear
  algebra.
\newblock {\em Bulletin of the American Mathematical Sociey\/}~{\em 59\/}(1),
  31--58.
\newblock https://doi.org/10.1090/bull/1722.

\bibitem[\protect\citeauthoryear{Freeman and Durden}{Freeman and
  Durden}{1998}]{freemandurden1998}
Freeman, A. and S.~L. Durden (1998, May).
\newblock A three-component scattering model for polarimetric {SAR} data.
\newblock {\em IEEE Transactions on Geoscience and Remote Sensing\/}~{\em
  36\/}(3), 963--973.
\newblock https://doi.org/10.1109/36.673687.

\bibitem[\protect\citeauthoryear{Hensley}{Hensley}{2021}]{hensley2021}
Hensley, S. (2021, Jul.).
\newblock The eigenvector-eigenvalue identity and radar polarimetry.
\newblock In {\em IEEE International Geoscience and Remote Sensing Symposium
  (IGARSS)}, Brussels, Belgium, pp.\  2--5.
\newblock https://doi.org/10.1109/IGARSS47720.2021.9554916.

\bibitem[\protect\citeauthoryear{Lee, Grunes, Ainsworth, Du, Schuler, and
  Cloude}{Lee et~al.}{1999}]{jsleeetal1999}
Lee, J.-S., M.~R. Grunes, T.~L. Ainsworth, L.-J. Du, D.~L. Schuler, and S.~R.
  Cloude (1999, Sep.).
\newblock Unsupervised classification using polarimetric decomposition and the
  complex {Wishart} classifier.
\newblock {\em IEEE Transactions on Geoscience and Remote Sensing\/}~{\em
  37\/}(5), 2249--2258.
\newblock https://doi.org/10.1109/36.789621.

\bibitem[\protect\citeauthoryear{Lee, Schuler, Ainsworth, Krogager, Kasilingam,
  and Boerner}{Lee et~al.}{2002}]{jslee_etal2002}
Lee, J.-S., D.~L. Schuler, T.~L. Ainsworth, E.~Krogager, D.~Kasilingam, and
  W.-M. Boerner (2002, Jan.).
\newblock On the estimation of radar polarization orientation shifts induced by
  terrain slopes.
\newblock {\em IEEE Transactions on Geoscience and Remote Sensing\/}~{\em
  40\/}(1), 30--41.
\newblock https://doi.org/10.1109/36.981347.

\bibitem[\protect\citeauthoryear{Nielsen}{Nielsen}{2020}]{fastLoewner2019}
Nielsen, A.~A. (2020, Oct.).
\newblock Fast matrix based computation of eigenvalues and the {L}oewner order
  in pol{SAR} data.
\newblock {\em IEEE Geoscience and Remote Sensing Letters\/}~{\em 17\/}(10),
  1727--1731.
\newblock https://doi.org/10.1109/LGRS.2019.2952202 and
  https://www2.imm.dtu.dk/pubdb/p.php?7151.

\bibitem[\protect\citeauthoryear{Nielsen}{Nielsen}{2022}]{EEI2022}
Nielsen, A.~A. (2022, Apr.).
\newblock The eigenvector-eigenvalue identity applied to fast calculation of
  {polSAR} scattering characterization.
\newblock {\em IEEE Geoscience and Remote Sensing Letters (Early Access)\/}.

\bibitem[\protect\citeauthoryear{Reigber, Scheiber, ager, Prats-Iraola,
  Hajnsek, Jagdhuber, Papathanassiou, Nannini, Aguilera, Baumgartner, Horn,
  Nottensteiner, and Moreira}{Reigber et~al.}{2013}]{FSAR2013}
Reigber, A., R.~Scheiber, M.~J. ager, P.~Prats-Iraola, I.~Hajnsek,
  T.~Jagdhuber, K.~P. Papathanassiou, M.~Nannini, E.~Aguilera, S.~Baumgartner,
  R.~Horn, A.~Nottensteiner, and A.~Moreira (2013, Mar.).
\newblock Very-high resolution airborne synthetic aperture radar imaging:
  Signal processing and applications.
\newblock {\em Proceedings of the IEEE\/}~{\em 101\/}(3), 759--783.
\newblock https://doi.org/10.1109/JPROC.2012.2220511.

\bibitem[\protect\citeauthoryear{Uspensky}{Uspensky}{1948}]{uspensky1948}
Uspensky, J.~V. (1948).
\newblock {\em Theory of Equations}.
\newblock McGraw Hill.

\bibitem[\protect\citeauthoryear{van Zyl}{van Zyl}{1993}]{vanzyl1992}
van Zyl, J.~J. (1993).
\newblock Application of {C}loude's target decomposition theorem to
  polarimetric imaging radar data.
\newblock In {\em Proceedings of SPIE 1748, Radar Polarimetry}, pp.\  184--191.
\newblock https://doi.org/10.1117/12.140615, (Event San Diego, USA, 1992).

\bibitem[\protect\citeauthoryear{van Zyl, Arii, and Kim}{van Zyl
  et~al.}{2011}]{vanzyletal2011}
van Zyl, J.~J., M.~Arii, and Y.~Kim (2011, Sep.).
\newblock Model-based decomposition of polarimetric {SAR} covariance matrices
  constrained for nonnegative eigenvalues.
\newblock {\em IEEE Transactions on Geoscience and Remote Sensing\/}~{\em
  49\/}(9), 3452--3459.
\newblock https://doi.org/10.1109/TGRS.2011.2128325.

\bibitem[\protect\citeauthoryear{Yamaguchi, Sato, Boerner, Sato, and
  Yamada}{Yamaguchi et~al.}{2011}]{yamaguchi_etal2011}
Yamaguchi, Y., A.~Sato, W.-M. Boerner, R.~Sato, and H.~Yamada (2011, Jun.).
\newblock Four component scattering power decomposition with rotation of
  coherency matrix.
\newblock {\em IEEE Transactions on Geoscience and Remote Sensing\/}~{\em
  49\/}(6), 2251–2258.
\newblock https://doi.org/10.1109/TGRS.2010.2099124.

\end{thebibliography}

\end{document}